\documentclass[twocolumn]{article}
\usepackage[utf8]{inputenc}
\usepackage{amsmath}
\usepackage{graphicx}
\usepackage{booktabs}
\usepackage{multirow}
\usepackage{hyperref}

\title{Deep Convolutional Neural Networks for predicting highest priority functional group in organic molecules}
\author{Kunal Khatri, Vineet Mehta, Manish Narwaria, and Bhaskar Chaudhary \\ Dhirubhai Ambani Institute of Information \& Communication Technology}
\date{}

\begin{document}

\maketitle

\begin{abstract}
Our work addresses the problem of predicting the highest priority functional group present in an organic molecule. Functional Groups are groups of bound atoms that determine the physical and chemical properties of organic molecules. In the presence of multiple functional groups, the dominant functional group determines the compound's properties. Fourier-transform Infrared spectroscopy (FTIR) is a commonly used spectroscopic method for identifying the presence or absence of functional groups within a compound. We propose the use of a Deep Convolutional Neural Networks (CNN) to predict the highest priority functional group from the Fourier-transform infrared spectrum (FTIR) of the organic molecule. We have compared our model with other previously applied Machine Learning (ML) method Support Vector Machine (SVM) and reasoned why CNN outperforms it.
\end{abstract}

\section{INTRODUCTION}
In organic chemistry, functional groups are specific groups of bound atoms which appear together within molecules, and determine the chemical and physical properties of the compounds \cite{1}. Precise identification of functional groups has significant applications in several fields such as biochemistry, molecular biology, medicinal chemistry, toxicity assessment, drug discovery, pharmaceuticals, and chemical nomenclature \cite{22}. Fourier-transform infrared (FTIR) spectroscopy is an important, commonly used spectroscopic method for identifying the presence or absence of functional groups within a compound \cite{12,20}. It is based on the interaction of infrared light with molecules present in a sample and the absorption of particular frequencies of the IR radiation. Characteristic absorption (or transmittance) patterns in the IR spectrum of a sample molecule helps in deducing the presence or absence of a specific functional group in the sample. 

Molecules having similar functional group exhibit similar chemical behaviour, but in the presence of multiple functional groups, the properties of the compound are determined by the dominant functional group present in it \cite{7}. When multiple functional groups are present, there is an overlap of the group frequencies of different functional groups \cite{20}, due to which patterns in the FTIR spectrum overlap and so it becomes difficult for a spectroscopist to predict the functional group. Machine learning has been applied to solve various problems in the field of chemoinformatics \cite{24,21,19}. Recently, Deep Learning has emerged and has been applied to fields of speech analysis \cite{16}, music analysis \cite{9} and sentence classification \cite{8}. In this paper, we have used deep Convolutional Neural Network (CNN) to tackle the problem of predicting the highest priority functional group present in an organic molecule. 

The main contribution of this paper are as follow: a) Collection of a large amount of FTIR spectrums of organic compounds. b) Applying deep learning on FTIR spectrums of organic compounds and predicting the highest priority functional group in them. 

The paper is divided into various sections. Section 2 discusses the related work on the problem and Section 3 describes the steps involved in data collection and preparation. Section 4 describes our CNN model. Lastly, Section 5 mentions the experiments conducted and the results obtained, followed by the conclusion in Section 6.

\section{RELATED WORK}
Since the 1990s, researchers have been addressing the problem of identification of Functional Groups. Robb et al. \cite{17,3} approached this problem using Artificial Neural Network (ANN). They used a single layer architecture of ANN to classify the molecules to learn and classify molecules which resulted in low accuracy (60\% on test data). Later, Fessenden et al. \cite{2} developed a model which also used a single hidden layer architecture and performed their experiment on 6 major functional groups. However, this was insufficient because a lot of important functional groups were missing and even the priority of functional groups was not taken into account. 

Over the next few years, a lot of studies \cite{5,6,4,18,25,26,21} were conducted by varying the number of input points, number of spectral features and the number of classes. However, all these papers that lacked the vital information of priority order of functional groups which helps in determining the name and chemical properties of the compound if it contains multiple functional groups. In our previous paper, Rajdeep and Rishikesh \cite{11} used two approaches to solve this problem from FTIR spectrum - intermediate approach, wherein they manually selected 4 features for each of the bond ranges and the fully automated approach, in which, they uniformly sample from the FTIR spectrum and obtain 250 features that they feed into their ML algorithm. We compare our results with theirs, using them as a benchmark. 

From the literature review, most of the work considered the presence/absence of functional groups or bonds and other structural features in the FTIR spectra. Furthermore, a lot of them did not consider functional group priority and only used ANNs with a single hidden layer. We aim to identify the highest priority functional group, using deep learning, from the FTIR spectrum of the molecule.

\section{DATA COLLECTION AND PREPARATION}
Spectral Database for Organic Compounds (SDBS) \cite{10} is a large open-source database with spectra of over 50000 compounds, where the data is in the form of FTIR images. Over a period of time, we collected the data of raw FTIR spectrum (Figure 1) of various functional groups to act as our dataset. 

\begin{figure}[h]
    \centering
    \caption{Raw data of FTIR spectrum, taken from SDBS \cite{10} database. The x-axis corresponds to the wave-numbers $(cm^{-1})$ and the y-axis corresponds to the transmittence(\%) for each of the x-values.}
\end{figure}

The FTIR graph is in the form of a $393\times320$ pixel image (Figure 1). The x-axis corresponds to the wave-numbers $(cm^{-1})$ and the y-axis corresponds to the transmittence(\%) for each of the x-values. Even though the wave-numbers on x-axis range from 4000 $cm^{-1}$ to $400~cm^{-1}$, since the image size is only $393\times320$ pixel, we need to map the points on x-axis only to 393 pixels. Hence, we cannot obtain the transmittence value for each wave number. So the obtained data in image form is further processed and quantified. Data cleaning, proper x-y mapping, scaling and interpolation is performed to prepare the data. The region from 1400 $cm^{-1}$ to $400~cm^{-1}$ is known as the fingerprint region and is unique to each organic compound so it is not taken into account while extracting transmittance values. Therefore, we extract the values of transmittance at each frequency from the region 4000 $cm^{-1}$ to 1400 $cm^{-1}$ and instead of taking all the 2600 points, we uniformly sample 404 points because the broad pattern of spectrum remains same even for lesser number of points. Finally, we normalize it to the range of [1,100]. Our final processed data is a sequence of 404 equidistant sampled points from the FTIR spectra (1D array), which is used as input data to all the ML models.

\section{CONVOLUTIONAL NEURAL NETWORK}
Convolutional neural networks \cite{15} are a specialized kind of neural network for processing data that has a known grid-like topology. Examples include time-series data and image data \cite{13}. Although primarily used in visual recognition contexts \cite{14}, convolutional architectures have been also successfully applied to unconventional data like speech \cite{16}, music analysis \cite{9}, Sentence classification \cite{8}. These efforts have shown that approaches taking advantage of data locality can provide viable solutions to problems encountered in other domains. Thus, we apply a deep Convolutional Neural Network(CNN) model to predict the highest priority functional group present in the organic molecule.

\begin{itemize}
    \item \textbf{Convolutional layer}: They comprise of a series of filters or learnable kernels which aim at extracting local features from the input, and each kernel is used to calculate a feature map or kernel map. Filters are tuned to the training set using training and backpropagation methods.
    \item \textbf{RELU Activation layer}: Role of the activation function/layer in a neural network is to produce a non-linear decision boundary via linear combinations of the weighted inputs. There are many non-linear activation functions like sigmoid, tanh. We have used Rectified Linear Units (ReLUs), which use the following activation function: $f(x)=max(0,x)$
    \item \textbf{MaxPooling layer}: Pooling layer basically downsamples the output(pooling operation) after the convolutional layer and activation layer.
\end{itemize}

For regularization and to prevent overfitting while training we use Dropout \cite{27} and Batch-Normalization \cite{23} layers.

\section{EVALUATIONS}
To ensure the efficiency of our method, we evaluated it on our dataset and compared it with the performance of previously applied method (SVM) on our dataset. We have total data of 4730 molecules and our data is skewed imbalanced class data, Table I.

\begin{table}[h]
\centering
\begin{tabular}{ll}
\toprule
Functional Group & Number of Data samples \\ \midrule
Carboxylic & 537 \\
Ester & 333 \\
Amide & 948 \\
Nitrile & 371 \\
Aldehyde & 357 \\
Ketone & 183 \\
Alcohol & 277 \\
Amine & 952 \\
Aromatic & 100 \\
Alkene & 100 \\
Alkyne & 49 \\
Alkane & 100 \\
Ether & 228 \\
Nitro & 159 \\ \midrule
Total & 4730 \\ \bottomrule
\end{tabular}
\caption{The class frequencies of highest priority functional groups present in our dataset.}
\end{table}

To compare the efficiency of our model (CNN) with the previously applied model (SVM), we evaluated both these models on A) 1600 molecules randomly sampled from our dataset, B) 3200 molecules randomly sampled from our dataset, and C) our entire dataset of 4730 molecules. We performed Stratified 10-folds cross validation test, wherein, the data randomly gets divided into 10 folds and out of these 10 folds, 9 folds are used as training data and 1 fold is used as testing data. The stratification ensures that each fold is representative of all strata of the data. This is performed 10 times, with different random seeds. The results of this experiment are tabulated in Table II.

\begin{table*}[h]
\centering
\begin{tabular}{lcccccc}
\toprule
\multirow{2}{*}{No.} & \multicolumn{2}{c}{1600 (A)} & \multicolumn{2}{c}{3200 (B)} & \multicolumn{2}{c}{4730 (C)} \\ \cmidrule(lr){2-3} \cmidrule(lr){4-5} \cmidrule(lr){6-7}
 & SVM & CNN & SVM & CNN & SVM & CNN \\ \midrule
1 & 64.51 & 68.85 & 66.18 & 75.01 & 69.12 & 79.70 \\
2 & 63.31 & 69.53 & 66.54 & 74.20 & 68.55 & 77.77 \\
3 & 64.40 & 67.32 & 65.93 & 74.51 & 68.30 & 76.89 \\
4 & 63.56 & 70.49 & 67.45 & 73.89 & 70.63 & 79.84 \\
5 & 63.87 & 68.14 & 67.62 & 73.97 & 78.80 & 68.68 \\
6 & 63.78 & 66.52 & 66.42 & 74.57 & 67.92 & 79.28 \\
7 & 64.63 & 68.77 & 66.62 & 74.76 & 69.20 & 77.12 \\
8 & 65.22 & 68.46 & 75.09 & 67.43 & 68.83 & 78.31 \\
9 & 63.82 & 67.61 & 67.49 & 73.90 & 67.22 & 77.80 \\
10 & 68.50 & 63.62 & 65.94 & 74.18 & 69.51 & 78.93 \\ \midrule
Mean & 64.07 & 68.42 & 66.76 & 74.40 & 68.79 & 78.43 \\ \bottomrule
\end{tabular}
\caption{Accuracy of the ML models on (A) 1600 molecules randomly sampled from our dataset, (B) 3200 molecules randomly sampled from our dataset and (C) entire dataset of 4730 molecules.}
\end{table*}

We also measure the Top-K score (accuracy), that is the percentage of predictions in which the actual highest priority functional group is among the top k predictions (based on probability) made by the ML model. This can be used for practical purpose. The results are tabulated in Table III.

\begin{table}[h]
\centering
\begin{tabular}{ccc}
\toprule
K & SVM & CNN \\ \midrule
1 & 68.42 & 78.43 \\
2 & 80.1  & 88.20 \\
3 & 89.2  & 94.3  \\ \bottomrule
\end{tabular}
\caption{Accuracy for Top K predictions made by ML models on the dataset, that is the percentage of predictions in which the actual highest priority functional group is among the top k predictions (based on probability) made by the ML model.}
\end{table}

\section{CONCLUSION}
We observed that Deep Convolutional Neural Networks perform significantly better than SVM. CNN takes advantage of data locality of the FTIR spectra sequence. Compared to SVM, CNN definitely requires more computation cost which it uses to extract neighbourhood/locality features from the spectra. We can therefore infer that locality pattern/features are more important to classify and identify highest priority functional group from FTIR spectra than mere values of transmittance at different frequencies.

\section{OTHER EVALUATIONS}
We perform random undersampling, i.e, we randomly sample 200 examples of each class and form a dataset on which we perform Stratified 10 fold test. To ensure as much randomness as possible, we repeat the process of random undersampling followed by evaluation (training and testing) on this generated dataset 10 times, each with different random seed values. 

\begin{table}[h]
\centering
\begin{tabular}{ccc}
\toprule
No. & SVM & CNN \\ \midrule
1 & 61.38 & 72.20 \\
2 & 61.45 & 71.59 \\
3 & 65.08 & 73.09 \\
4 & 61.01 & 71.83 \\
5 & 61.59 & 71.38 \\
6 & 62.05 & 71.46 \\
7 & 61.30 & 72.34 \\
8 & 62.78 & 71.75 \\
9 & 63.75 & 72.94 \\
10 & 61.67 & 72.50 \\ \midrule
Mean & 62.20 & 72.11 \\ \bottomrule
\end{tabular}
\caption{Accuracy of ML models on undersampled data, i.e, randomly sample 200 examples of each class and form a dataset on which we perform Stratified 10 fold test.}
\end{table}

Next, we apply our model on the dataset used by the previous paper \cite{11} which is a subset of the dataset used by us, containing 1380 molecules.

\begin{table}[h]
\centering
\begin{tabular}{ccc}
\toprule
SVM & RF & CNN \\ \midrule
85.96 & 86.02 & 90.32 \\ \bottomrule
\end{tabular}
\caption{Accuracy on applying ML models on previous paper's \cite{11} dataset.}
\end{table}

We design an experiment to investigate the reason for this drastic difference in accuracy. For convenience we call the data used by them as Data-old. From the experiment our goal is to assert that the data used was more specific, not generalizable or that it contained less global patterns. In this experiment, we compare two methods:
1. Train ML model on Data-old and Test it on same amount of Random Undersampled data that we have collected, i.e Data-new
2. Do the opposite, Train ML model on Data-new and test it on Data-old

\begin{table}[h]
\centering
\begin{tabular}{lcc}
\toprule
Training Data & SVM & CNN \\ \midrule
DATA-NEW & 68\% & 74\% \\
DATA-OLD & 50\% & 59\% \\ \bottomrule
\end{tabular}
\caption{Comparing accuracy of ML models, trained on different data.}
\end{table}

The reasoning behind this experiment is that, if the test set accuracy of ML model is similar then we can infer that the model is able to learn similar important features from data and that the generalizability of data is similar. But if the test set accuracy of ML model is significantly higher when training on Data-new than on other, we can infer that the model is able to generalize or comparably learn better from Data-new.

\bibliographystyle{unsrt}
\bibliography{myBib_arxiv}

\end{document}